\documentclass{article}
\usepackage{graphicx}
\usepackage{booktabs}
\usepackage{amssymb}
\usepackage{multirow}
\usepackage{float}
\usepackage{subcaption}
\usepackage{placeins}
\usepackage[preprint]{corl_2026} 

\title{X-Morph: Human Motion Priors for Scalable Robot Learning Across Morphologies}

\author{
\textbf{Ritwik Sharma}$^{*}$$^{\dagger}$\quad
\textbf{Shivam Sood}$^{*}$ \quad
\textbf{Arhaan Jain}\\[0.35em]
\textbf{Shyam Charan Kesavamoorthi} \quad
\textbf{Chengyang He} \quad
\textbf{Guillaume Adrien Sartoretti}\\[0.35em]
National University of Singapore\\[0.25em]
{\small $^{*}$Equal contribution \quad $^{\dagger}$Corresponding author}
}

\begin{document}
\maketitle


\begin{abstract}

Recent progress in humanoid behavior models has been driven in large part by abundant human motion data, but comparable motion data is scarce for non-humanoid legged robots such as quadrupeds, hexapods, and quadruped manipulators.
A promising alternative is to repurpose human motion across embodiments; however, direct retargeting often produces motions that are visually plausible yet physically inconsistent or difficult to track under robot dynamics.
We present X-Morph, a human-motion-to-robot-behavior pipeline that converts human motion into deployable locomotion and loco-manipulation policies for diverse non-humanoid legged morphologies.
A cross-morphology retargeting stage converts human motions into kinematically plausible, intent-preserving robot references, which are then tracked by a privileged RL policy and distilled into a causal student policy.
We evaluate X-Morph on three morphologically distinct platforms: a quadruped, a hexapod, and a quadruped equipped with a manipulator.
The resulting policies track diverse retargeted motions, generalize to unseen human motions, and support downstream use cases including video-based teleoperation, behavior-prior control, and text-conditioned motion generation.
These results suggest that large-scale human motion can serve as a substrate for learning broad, reusable behavior priors beyond humanoid robots. Project page: \href{https://maker-rat.github.io/morph/}{https://maker-rat.github.io/morph/}
\end{abstract}

\keywords{Cross-Morphology Motion Transfer, Human Motion Retargeting, Legged Robot Learning} 


\section{Introduction}

Humanoid robot control has advanced rapidly, driven by large-scale motion datasets and motion-prior learning methods that enable whole-body behaviors without extensive task-specific engineering~\citep{Peng_2018, sreenath2025beyondmimicmotiontracking, weng2025hdmi, luo2026sonicsupersizingmotiontracking, rempe2026kimodoscalingcontrollablehuman}.
In contrast, non-humanoid legged robots, such as quadrupeds, hexapods, and quadruped manipulators, lack comparably rich behavior datasets.
As a result, skill acquisition for these systems remains dependent on robot-specific demonstrations, hand-crafted rewards, or small curated skill libraries, forcing each new morphology to rebuild much of its behavior data from scratch.

A natural question is whether abundant human motion can serve as a reusable behavior substrate for these non-humanoid legged robots.
While human motions cannot directly dictate how a quadruped or hexapod should move, they contain transferable features: contact timing, body coordination, limb sequencing, task intent, and expressive whole-body patterns.
Prior work has shown that meaningful correspondences can be learned between human motion and non-human or non-humanoid embodiments, including legged robots~\citep{li2024crossloco, kim2025moreflowmotionretargetinglearning}.
Computer animation has similarly shown that motion can be transferred or generated across characters with different skeletal topologies~\citep{zhang2025motion2motioncrosstopologymotion, bermano2025anytopcharacteranimation}.
Together, these results suggest that cross-morphology retargeting is a promising route toward reusable behavior priors.
However, retargeting quality alone is not sufficient: to serve as a robot behavior prior, the transferred motion must be physically consistent, trackable under closed-loop control, and usable through practical interfaces such as video, language, and downstream task learning.
The central challenge is that cross-topology retargeting typically produces only a candidate reference. 
Even visually plausible transfers may contain foot skating, ground penetration, floating contacts, joint-limit violations, or dynamically inconsistent transitions. For robots, these artifacts matter because the reference must be trackable under actuation limits, contact dynamics, noisy state estimates, and closed-loop control. 
Thus, the objective is not literal human imitation, but the preservation of behaviorally relevant structure while allowing morphology-specific deviations required for physical execution.

We propose X-Morph, a cross-morphology behavior-transfer system that converts human motion into executable skills for diverse legged morphologies. 
X-Morph treats retargeting as an intermediate reference-generation step rather than the final output: human motions are first mapped to target robot references, then corrected for contact and kinematic consistency, tracked by a privileged reinforcement learning (RL) policy, and finally distilled into deployable causal student policies. 

The retargeting stage maps human motion intent to a target robot morphology, the correction stage improves contact and ground-interaction consistency, and the tracking stage converts the corrected references into closed-loop policies.
We study the complete path from human motion to deployable non-humanoid robot behavior.
This lets us evaluate whether transferred human motion can function as an interactive behavior prior across distinct morphologies, supporting live video teleoperation, live text-to-skill execution, and downstream task learning.
The key contributions of our paper are summarized below:
\begin{enumerate}
    \item We present an end-to-end system that connects human-to-non-humanoid retargeting with physics-aware reference correction, privileged tracking, and causal policy distillation for deployable legged-robot behavior.
    \item We evaluate the resulting pipeline across three morphologically distinct legged platforms: a quadruped, a hexapod, and a quadruped manipulator, measuring tracking quality, physical consistency, unseen-motion generalization, and downstream usability.
    \item We demonstrate that the learned motion library can act as an interactive behavior prior, enabling live video teleoperation and text-to-skill execution from transferred human motion.
\end{enumerate}

\section{Related Work}
\label{sec:litreview}

\textbf{Human motion priors for humanoid control:}
Human motion has become a powerful supervision signal for physics-based character control and humanoid robot learning. Early systems such as DeepMimic~\citep{Peng_2018} show that motion clips can be converted into robust controllers through reinforcement learning, while later methods learn reusable motion priors or skill spaces from large unstructured datasets~\citep{Peng2021AMP, peng2022ase, luo2024universal}. Recent humanoid systems scale this idea through general motion tracking, privileged imitation, and generative motion priors, enabling increasingly diverse whole-body behaviors from large motion corpora~\citep{luo2023perpetual, chen2025gmt, sreenath2025beyondmimicmotiontracking, luo2026sonicsupersizingmotiontracking, mu2025smp}. These works motivate motion data as a reusable behavior prior, but primarily operate on humanoid or near-humanoid embodiments where the source and target share similar body structure. For quadrupeds, animal-motion imitation has similarly shown that reference motion can reduce manual reward design and produce agile real-robot behaviors~\citep{peng2020learning}. This motivates our question: can human motion serve as a reusable behavior prior for robots with substantially different morphology and contact structure?

\textbf{Cross-morphology retargeting and robot control:}
Motion retargeting across characters has been widely studied in graphics and animation, where the goal is typically to preserve semantic content, style, or visual plausibility across different skeletons~\citep{Aberman_2020, zhao2023posetomotioncrossdomainmotionretargeting, chen2025motion2motion, bermano2025anytopcharacteranimation}. Recent methods further address larger topology changes through encoder-decoder models, part-based representations, textual conditioning, or spatial correspondences~\citep{zhang2025motion2motioncrosstopologymotion, liu2026palumpartbasedattentionlearning, Hu_2024}. Closer to robotics, HumanConQuad retargets human motion to quadruped motion and trains imitation policies for human-motion control of quadrupeds~\citep{kim2022humanconquad}. ACE learns adversarial correspondence embeddings for retargeting human motion to nonhuman characters, including robots with substantially different body structures~\citep{li2023ace}. CrossLoco jointly learns human-robot correspondences and legged skills through guided reinforcement learning~\citep{li2024crossloco}. STMR combines spatial retargeting, temporal refinement, and imitation learning to obtain deployable quadruped motions~\citep{yoon2025spatio}. MoReFlow formulates cross-morphology motion retargeting as flow matching between motion distributions~\citep{kim2025moreflowmotionretargetinglearning}. Together, these works establish that meaningful human-to-robot or cross-species correspondences can be learned, and that several forms of physical feasibility and deployment can be incorporated into cross-morphology motion transfer. 
X-Morph builds on this progress, but differs in treating retargeted motion not as the final output but as an intermediate robot-learning representation that is corrected, tracked, distilled, and reused through interactive interfaces.
We evaluate this complete path across a quadruped, a hexapod, and a quadruped manipulator.
\begin{figure}
    \centering
    \includegraphics[width=1\linewidth]{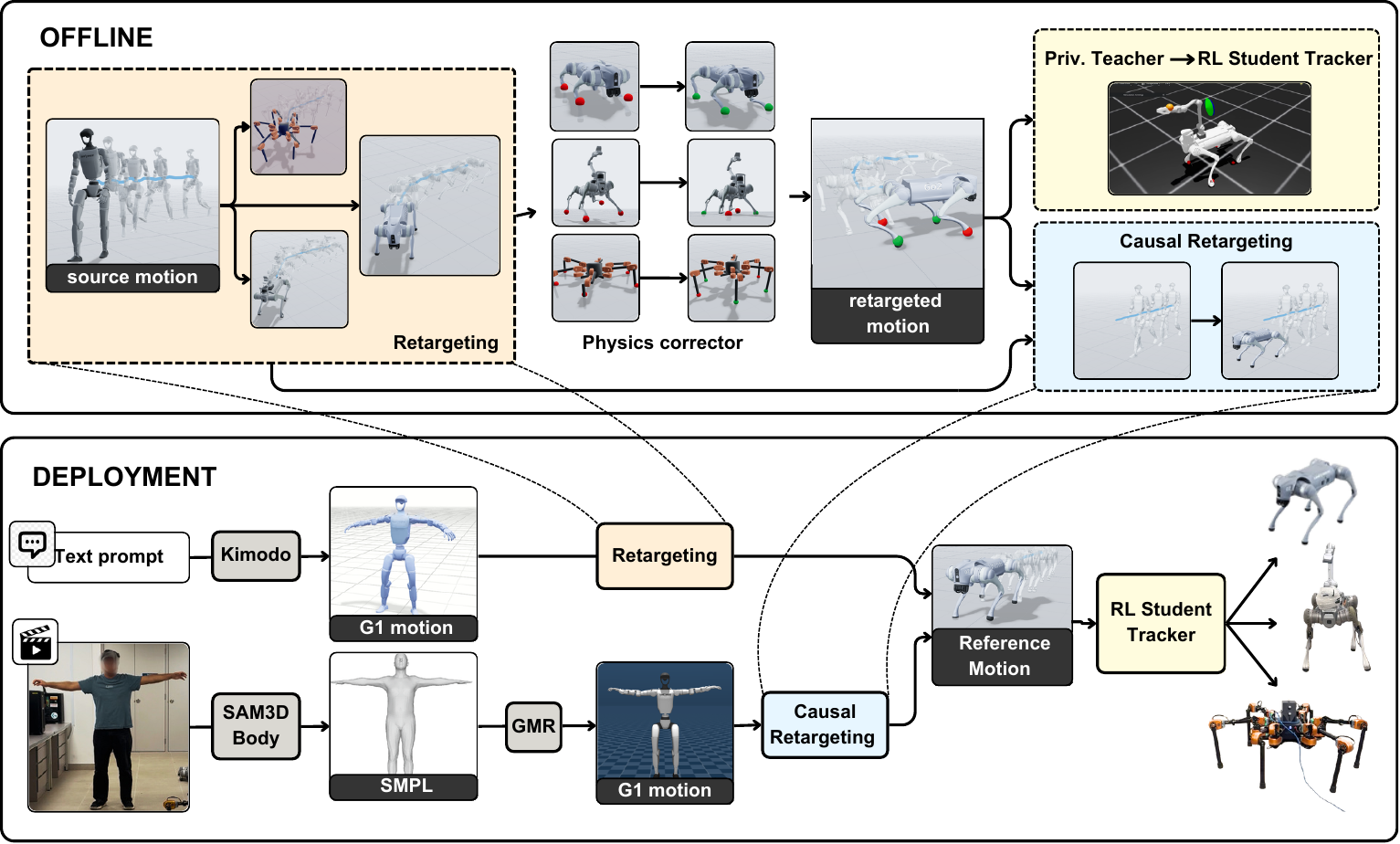}
    \vspace{-0.5cm}
    \caption{\textbf{X-Morph framework.} Source human/G1 motions are converted into target robot references by a cross-morphology retargeting model and then refined by a physics-aware corrector to reduce contact and ground-interaction artifacts. The resulting clean retargeted motions provide reference data for learning the reference-conditioned tracker and for distilling a causal retargeting model from the offline retargeting model. At deployment, X-Morph supports two separate workflows. In the text-conditioned workflow, a text prompt is converted into a G1 motion through Kimodo; this motion is then retargeted as a whole into a robot reference. In the video-driven workflow, an RGB video stream is converted to SMPL using FastSAM3D Body~\citep{yang2026fastsam3dbody}, retargeted to G1 motion using GMR~\citep{araujo2025retargeting}, and then mapped online to a robot reference by the causal retargeting model. Both workflows produce the same reference-motion interface, which is executed by the deployed RL student tracker across multiple non-humanoid morphologies.}
    \label{fig:framework}
\end{figure}

\section{X-Morph Framework}
\label{sec:methodology}
Fig.~\ref{fig:framework} summarizes X-Morph. The framework consists of three learned modules: an offline retargeting network, a reference-conditioned tracking policy, and a causal retargeting network for interactive deployment. The following sections describe the reference-generation problem, the morphology-aware retargeter, physics-aware correction, tracker training, and causal deployment interfaces.

\subsection{Problem Formulation}
Our goal is to convert a source human motion sequence $\mathbf{x}^h = \{x^h_t\}_{t=1}^{T}$ into an executable behavior for a target robot morphology $\mathcal{M}^r$, while preserving the source motion's task intent, root movement, and task-relevant body or end-effector coordination. 
We represent a morphology by its kinematic tree and additional metadata such as joint limits,
nominal base height, and feet or end-effectors. Each source-target pair is also
associated with a retargeting specification $\mathcal{C}^{h \rightarrow r}$,
which defines semantic body-part correspondences and task-relevant contact or end-effector mappings. 
We decompose this problem into reference generation and reference tracking. A cross-morphology retargeting model first maps the source motion to a target
robot reference trajectory,
\[
    \hat{\mathbf{s}}^r =
    f_{\theta}(\mathbf{x}^h, \mathcal{M}^r, \mathcal{C}^{h \rightarrow r}),
\]
where each $\hat{s}^r_t$ contains target joint positions and root motion
features; contact and end-effector quantities used by losses or downstream
tracking are derived from this trajectory through the robot kinematic model. This reference may then be
refined by a physics-aware corrector,
\[
    \tilde{\mathbf{s}}^r =
    c_{\psi}(\hat{\mathbf{s}}^r, \mathbf{x}^h, \mathcal{M}^r,
    \mathcal{C}^{h \rightarrow r}),
\]
to reduce artifacts such as foot skating, ground penetration, floating contacts,
and long-horizon root drift.
The corrected trajectory $\tilde{\mathbf{s}}^r$ is used as a reference for a
closed-loop robot policy. At execution time, the policy observes the robot state
$o_t$ and a short horizon of reference states $\tilde{s}^r_{t:t+H}$, and outputs
an action
\[
    a_t \sim \pi_{\phi}(a_t \mid o_t, \tilde{s}^r_{t:t+H}).
\]
Thus, in addition to synthesizing visually plausible target motion, the objective is also to obtain a reference-policy pair that can be physically executed on the robot.

\subsection{Morphology-Aware Reference Generation}
We build our retargeting network on the body-part retargeting framework
of~\citet{Hu_2024}. The key idea is to encode motion through semantically
corresponding body parts rather than requiring identical source and target
kinematic trees. In our setting, human motions from datasets such as AMASS and
LAFAN1 ~\citep{AMASS, harvey2020robust} are first represented in a Unitree G1 humanoid embodiment. This gives us a
robot-description-based source representation with known joint order, joint limits, root pose, and forward kinematics, making cross-morphology learning more structured than directly mapping from raw SMPL parameters to each target robot.
For a robot morphology with $n_r$ actuated joints, we represent each motion frame
as
\[
    m_t^r = [q_t^r,\; v_t^r,\; \omega^r_{z,t}]
    \in \mathbb{R}^{n_r + 4},
\]
where $q_t^r \in \mathbb{R}^{n_r}$ are joint angles in the MuJoCo XML joint
order and sign convention, $v_t^r \in \mathbb{R}^{3}$ is the root linear velocity in the local heading frame, and $\omega^r_{z,t} \in \mathbb{R}$ is the root yaw rate. The retargeting network predicts this normalized motion representation for the target morphology, and the target root trajectory is reconstructed by integrating the predicted local velocity and yaw rate.

Following \citet{Hu_2024}, the model contains morphology-specific motion encoders and decoders connected through a shared body-part latent space. A pose-aware attention module pools joint-level features into body-part tokens, which are then decoded using the target morphology. We train the retargeting network with the standard PAN-style reconstruction, cycle-consistency, adversarial, and FK-based kinematic losses, which encourage each morphology's autoencoder to reconstruct its own
motions, keep transferred motions in a shared latent space, and preserve kinematic structure after cyclic retargeting.

For our robot-learning setting, we augment this objective with terms that improve trackability. Physics losses penalize foot skating, floating, and ground penetration using target-robot forward kinematics and nominal ground height, while manipulation losses align task-relevant end-effectors according to $\mathcal{C}^{h \rightarrow r}$. These terms bias the learning toward references that are not only kinematically plausible, but useful for downstream tracking and interaction.

\subsection{Physics-Aware Reference Correction}
The retargeting network produces a morphology-consistent reference, but this reference is not guaranteed to be physically useful for robot tracking. In practice, transferred motions may contain contact artifacts such as foot skating, floating feet, ground penetration, discontinuous root motion, or end-effector drift. We therefore apply an offline physics-aware corrector to the retargeting network output before using the motion for policy training.

The corrector is a temporal residual model that operates on full motion clips and edits joint angles and root trajectory components while preserving the retargeted motion's local root motion. Given the network output $\hat{\mathbf{s}}^r$, the corrector predicts a cleaned trajectory $\tilde{\mathbf{s}}^r$ that reduces contact and kinematic artifacts while maintaining the source motion intent. The resulting references are used for tracker training, student distillation, causal retargeting, and downstream deployment.

\subsection{Reference-Conditioned Policy Tracking}
\label{sec:motion_tracking}
We train a reference-conditioned tracker that converts cleaned robot references into closed-loop control policies. At each step, the tracker observes the robot state $o_t$ and a short horizon of reference features $\tilde{s}^r_{t:t+H}$, including target joint positions, base pose and velocity features, and foot or end-effector states: $ a_t \sim \pi_\theta(a_t \mid o_t, \tilde{s}^r_{t:t+H})$

\textbf{Privileged teacher:} To obtain a strong tracking expert, we first train a privileged teacher with clean full-state observations. The teacher receives deployable proprioception augmented with privileged full-state information, including extended reference context, body poses, base-height error, root velocities, joint torques, contact state, and motion identity. We train the teacher with APEX-style action priors~\citep{sood2025apexactionpriorsenable} and DeepMimic-style~\citep{Peng_2018} tracking rewards.

\textbf{Student distillation.}
The deployed tracker is a causal student distilled from the privileged teacher. The student observes only deployable proprioception and a compact reference stream consisting of the current and one-step-future reference features. Given a trained teacher $\pi_{\theta^\star}$, we train the student $\pi_\phi$ using supervised action distillation:
\begin{equation}
    \mathcal{L}_{\mathrm{distill}}(\phi)
    =
    \mathbb{E}_{t}
    \left[
    \left\|
    \pi_\phi(\tilde{o}_{t-k:t}, \tilde{s}_{t:t+1})
    -
    \pi_{\theta^\star}(o^{\mathrm{priv}}_t)
    \right\|_2^2
    \right],
\end{equation}
where tildes denote noisy deployable observations and reference inputs.

\subsection{Causal Retargeting for Interactive Deployment}
For interactive and real-time deployment, the offline retargeting-and-correction stack is computationally expensive and non-causal. We therefore train a causal retargeting student to predict the cleaned target references from recent source motion and autoregressive target context.
The student is trained to imitate the cleaned target references produced by the
offline retargeting network and physics corrector. For each paired training clip, the input is a
short causal history of source G1 motion features together with a short history
of the student's previous target predictions. The target is the corresponding
cleaned robot reference frame,
\[
    \bar{s}^r_t =
g_{\eta}\!\left(m^g_{t-k:t}, \bar{s}^r_{t-\ell:t-1}\right),
\]
where $\bar{s}^r_t$ denotes the causal student's predicted robot reference, $m^g_{t-k:t}$ denotes the recent G1 motion context and
$\bar{s}^r_{t-\ell:t-1}$ denotes autoregressive target context. The student
predicts the target robot joint positions and root motion features for the
current frame.

\textbf{Video input:} For live video teleoperation, we recover a human SMPL~\citep{10.1145/2816795.2818013} motion estimate from a monocular RGB stream and convert it online to the G1 humanoid representation using GMR~\citep{araujo2025retargeting}. This produces a stream of G1 motion features in the same format used by the retargeting network. The causal retargeting student then maps this stream to target robot references, which are executed by the deployed tracker.

\textbf{Text-Conditioned Skill Execution:}
As shown in the deployment branch of Fig.~\ref{fig:framework}, X-Morph can also use text-conditioned human-motion generation or retrieval as input. A text prompt is converted into a G1 motion sequence using a human-motion model, after which the same retargeting and tracking stack converts the generated motion into an executable robot behavior. This allows language commands to select or synthesize behaviors without training a separate language-conditioned controller for each target morphology.


\section{Experimental Results}
\label{sec:result}
We evaluate X-Morph along five axes: (i) cross-morphology motion execution, (ii) physical tracking of retargeted references, (iii) interactive control through live video teleoperation, (iv) live text-to-skill execution using generated or retrieved human motions, and (v) downstream task learning with transferred human-motion priors. We further ablate the major stages of the pipeline to test whether retargeting alone is sufficient for robust execution.

\subsection{Cross-Morphology Motion Execution}
We first test whether X-Morph can convert human motions into executable behaviors across a quadruped, a hexapod, and legged systems performing object-interaction behaviors. The goal is not literal joint-level imitation, but preservation of intent, timing, body coordination, and contact structure under each robot's morphology.

\paragraph{Video-Driven Teleoperation}
\begin{figure}[t]
    \centering
    \includegraphics[width=0.9\linewidth]{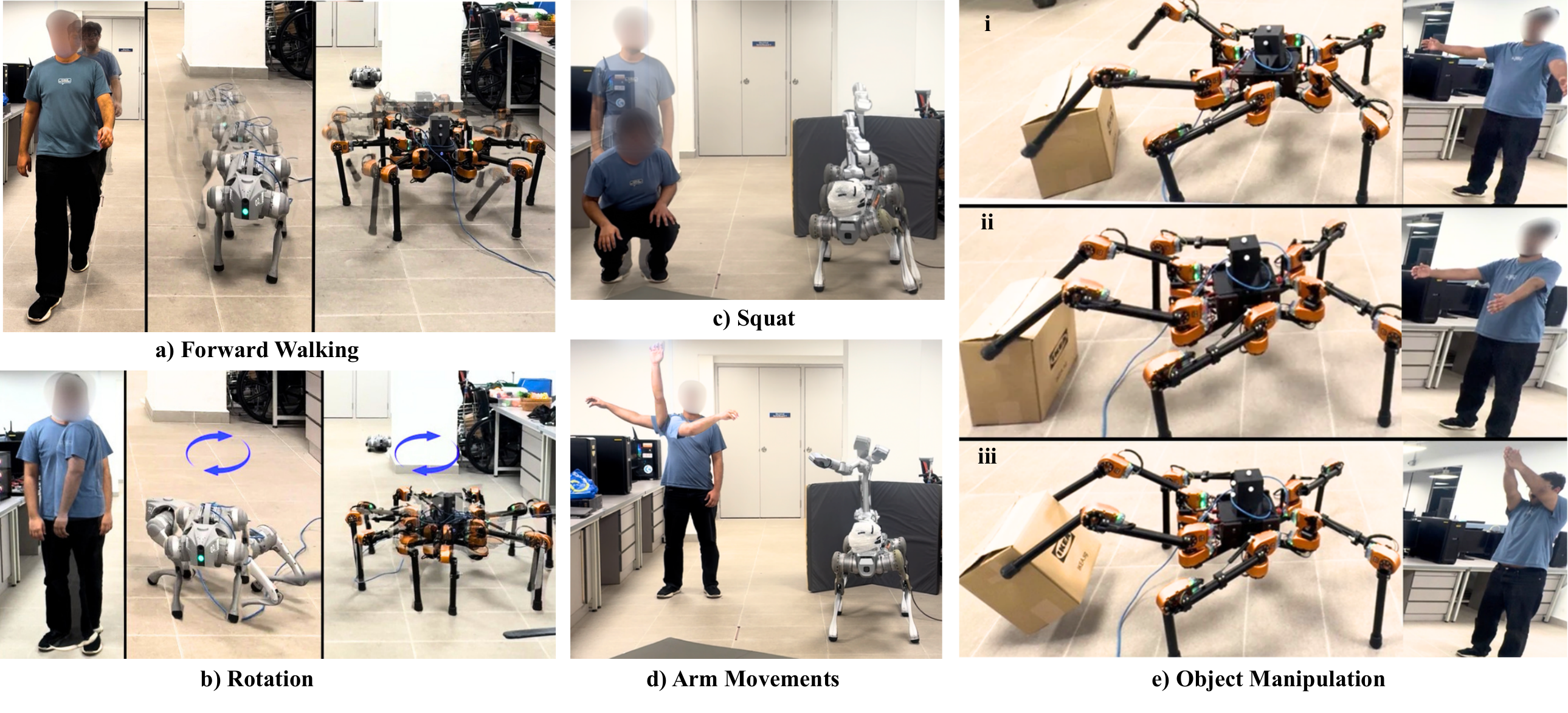}
    \hspace{0.04\linewidth}
    \caption{
    \textbf{Video-driven teleoperation across non-humanoid morphologies.}
    X-Morph converts monocular human motion into executable robot references for multiple target platforms. 
    (a) Forward walking is transferred to both a quadruped and a hexapod.
    (b) Human body rotation induces turning behaviors on both robots.
    (c) A squat motion is retargeted to a quadruped while preserving the high-level lowering intent.
    (d) Large arm motions are converted into expressive whole-body robot motions.
    (e.i--e.iii) Object-interaction motions are transferred to a hexapod, where different human arm poses produce corresponding whole-body reaching and box-interaction behaviors.
    }
    \label{fig:video_teleop}
\end{figure}
We evaluate X-Morph as a video-based teleoperation interface. Given monocular human video, we estimate a source motion stream, convert it to the G1 representation, retarget it online to the target morphology, and execute the resulting reference with the deployed tracker.  With visualization disabled, the live pipeline publishes retargeted robot
references at up to 28.9~Hz using a 30~Hz camera stream. Fig.~\ref{fig:video_teleop} shows that the same interface transfers walking, turning, squatting, expressive upper-body motion, and object-interaction intent across non-humanoid platforms. The box-picking example is not intended to demonstrate a high-success manipulation policy. Instead, it shows how X-Morph can convert human object-interaction motions into structured robot trials that are useful for downstream data collection. Although open-loop success remains limited due to accumulated perception, retargeting, and contact-timing errors, these trials provide task-relevant initialization data for training a closed-loop manipulation policy. As shown in the videos, diverse whole-body motions such as dancing and warm-up movements can also be transferred to non-humanoid robots.

\paragraph{Text-Conditioned Motion Execution}

Next, we evaluate whether the transferred motion library can be used as a language-driven behavior interface. A user provides a text command, which is used to retrieve or generate a corresponding human motion. X-Morph then retargets this motion to the target morphology and executes it through the same reference-conditioned tracker. This experiment tests a central advantage of our formulation: once human motion is converted into robot-trackable references, existing human-motion generation and retrieval interfaces can be reused for non-humanoid robot control.

\begin{figure}[h]
    \centering
    \includegraphics[width=0.75\linewidth]{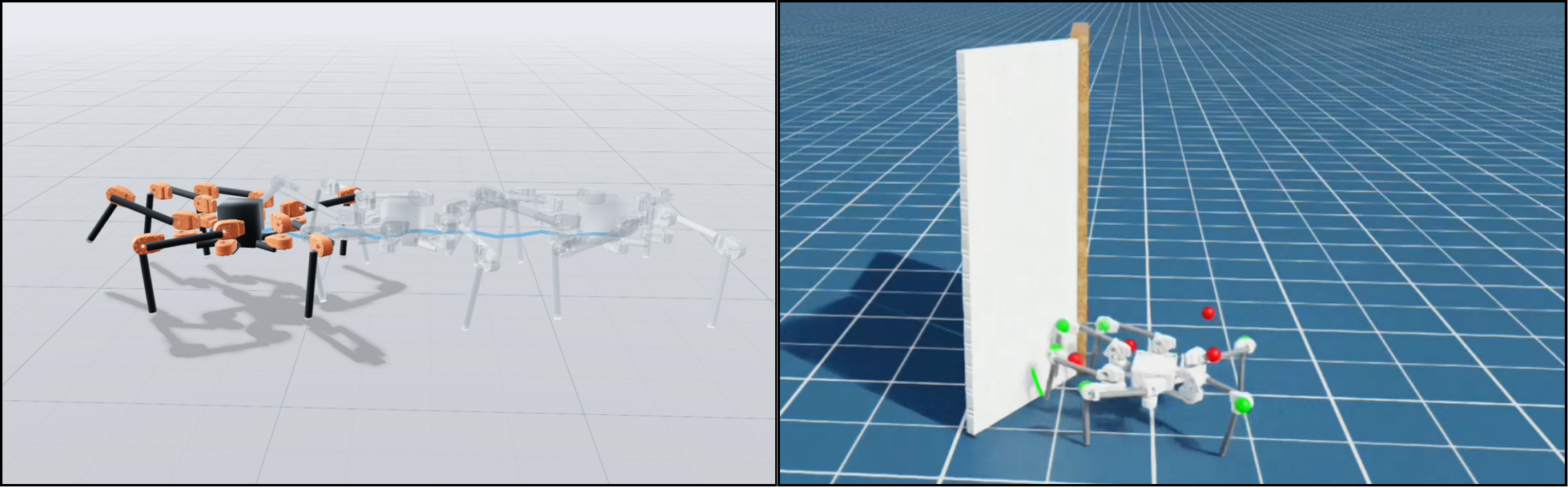}
    \caption{
    Downstream door-opening case study. Left: X-Morph retargets a
    Kimodo-generated human interaction into a hexapod loco-manipulation
    reference, producing coordinated body motion and front-leg reaching.
    Right: a downstream policy initialized from this structured prior executes
    a door-opening behavior in simulation. This result is intended as a
    qualitative demonstration of task-relevant initialization, not as a
    controlled comparison against reinforcement learning from scratch.
    }
    \label{fig:hexapod_door_opening}
\end{figure}

\paragraph{Retargeted Motions as Downstream Behavior Priors:}
Beyond direct motion execution, we show that retargeted human motions can provide useful behavioral structure for downstream robot learning.
Fig.~\ref{fig:hexapod_door_opening} depicts a qualitative downstream case study in which a loco-manipulation prior generated by X-Morph is used to initialize a hexapod door-opening behavior. The resulting policy learns a coordinated reaching and body-positioning strategy that can open the door in simulation. We do not claim a sample-efficiency improvement over reinforcement learning from scratch, since a controlled baseline comparison is left for future work.
This highlights a key use case of X-Morph: human motion need not be treated as a final controller.
Instead, it can serve as a reusable behavior prior that supplies diverse, morphology-specific initial behaviors for later task learning. This mirrors the role of large human motion datasets in humanoid control, but extends the idea to non-humanoid legged robots where comparable robot-specific behavior datasets are scarce.

\begin{figure}[!htbp]
    \centering
    \includegraphics[width=0.9\linewidth]{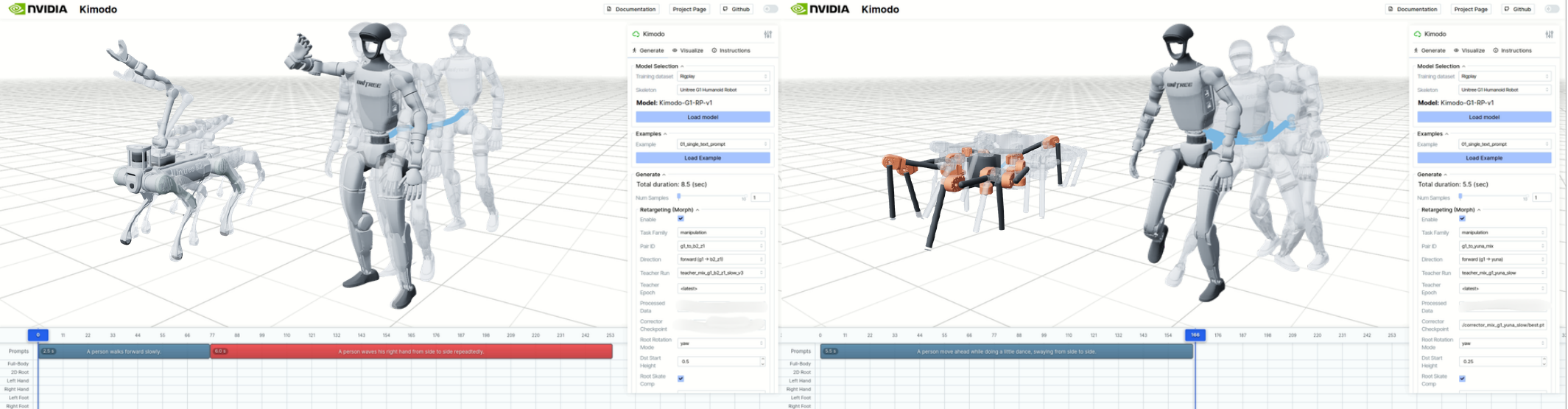}
    \caption{
    \textbf{Text-conditioned skill execution.}
    A language command is converted into a human motion through a text-conditioned human-motion model or retrieval system. X-Morph retargets the resulting G1 motion to the target morphology and executes it with the same deployed tracker.
    }
    \label{fig:kimodo}
\end{figure}

\subsection{Reference Quality and Generalization to Unseen Human Clips}

\begin{table}[t]
\centering
\caption{
Go2 retargeting-reference ablation on 33 matched locomotion clips.
The physics-aware corrector reduces contact artifacts while preserving the
overall motion scale. Slip is mean horizontal foot speed for feet near the
ground; penetration and floating are measured from FK foot heights.
}
\label{tab:go2_reference_ablation}
\resizebox{\linewidth}{!}{
\begin{tabular}{lcccccc}
\toprule
Method &
Root speed p95 &
Foot slip $\downarrow$ &
Penetration p95 $\downarrow$ &
Contact height err. $\downarrow$ &
Floating err. $\downarrow$ &
Joint acc. p95 $\downarrow$ \\
&
(m/s) & (cm/s) & (cm) & (cm) & (cm) & (rad/s$^2$) \\
\midrule
Raw retargeting network &
0.915 & 58.76 & 11.34 & 6.45 & 3.09 & 32.61 \\
Retargeting + physics corrector &
0.798 & 42.76 & 6.02 & 3.60 & 1.88 & 28.08 \\
\midrule
Relative change &
-- & \textbf{27.2\%} & \textbf{46.9\%} & \textbf{44.2\%} &
\textbf{39.3\%} & \textbf{13.9\%} \\
\bottomrule
\end{tabular}
}
\end{table}

\begin{table}[h]
\centering
\small
\caption{Yuna hexapod sim2sim tracking under live video reference streaming. Metrics are
computed over the active segment after removing startup and shutdown buffers.
Lower is better for all metrics.}
\label{tab:yuna_live_tracking}
\resizebox{\linewidth}{!}{
\begin{tabular}{lcccccc}
\toprule
Tracker reference &
Joint MAE &
Root vel. RMSE &
Yaw-rate RMSE &
Base height std. &
Foot slip mean \\
& (deg) & (m/s) & (rad/s) & (cm) & (cm/s)\\
\midrule
Uncorrected refs & 6.57 & 0.479 & 0.896 & 2.12 & 29.29 \\
Corrected refs & 5.45 & 0.413 & 0.651 & 1.72 & 24.30 \\
\midrule
Rel. Improvement & 17.4\% & 13.9\% & 27.5\% & 18.7\% & 17.5\% \\
\bottomrule
\end{tabular}
}
\end{table}

Table~\ref{tab:go2_reference_ablation} shows that retargeting alone produces plausible but physically noisy references. Adding the offline corrector reduces near-ground foot slip by 27.2\%, penetration by 46.9\%, contact-height error by 44.2\%, and low-speed floating by 39.3\%, while only moderately reducing the root-speed scale. 

Table~\ref{tab:yuna_live_tracking} shows that training the tracker on corrected  references improves overall tracking and base stability under the same live reference stream, with a 17.4\% reduction in joint tracking error and a 27.5\% reduction in yaw-rate error. Mean foot slip also decreases. This supports our design choice of treating retargeting as a candidate reference generator rather than the final motion, and the use of corrected references for training and tracking.

We also observe generalization on held-out human motions and motion variants not used for retargeter or tracker training. Qualitatively, the policies remain stable under larger limb excursions and side-specific interaction motions, suggesting that the learned representation captures reusable body-part structure rather than memorizing target joint trajectories. These results can be best seen in the attached video and Appendix.


\section{Conclusion}
\label{sec:conclusion}
We presented X-Morph, a system for converting human motion into executable behaviors for non-humanoid legged robots. These motions are used to produce robot-specific reference trajectories that are corrected, tracked with reinforcement learning, and distilled for realtime deployment. Across a quadruped, a hexapod, and a quadruped manipulator, the system shows that human motion can provide useful structure even when the target morphology differs substantially from the source.
Our results suggest a practical path for reusing human motion beyond humanoid robots. Retargeted motions can be executed directly, connected to video or text derived human motion sources, and used as behavior priors for downstream reinforcement learning. More broadly, this work points toward a way of using large human motion datasets as reusable behavioral scaffolds for a wider class of legged robots.
\section{Limitations}
\label{sec:limitations}
X-Morph does not guarantee that every human motion can be meaningfully transferred to every morphology. Currently the method relies on manually specified semantic correspondences between source and target body parts, and poor correspondences can produce references that are difficult to correct or track. Our physics-aware corrector reduces common artifacts such as foot skating and ground penetration, but it is not a full trajectory optimizer and does not guarantee dynamic feasibility. The current experiments focus primarily on flat or moderately structured terrain, and more complex terrain may require terrain-aware correction and tracking. Finally, video-driven deployment depends on the quality and latency of monocular human pose estimation, which can introduce errors under occlusion, fast motion, or unusual camera viewpoints.

\section{Acknowledgments}
This research was supported by the Singapore Ministry of Education (MOE), as well as by NUS under their Robotics Grand Challenge.

\clearpage
\bibliography{bibliography}  

\clearpage
\appendix
\begin{center}
{\LARGE\textbf{Appendix}}
\end{center}
\vspace{0.5em}
\section{Retargeting Specifications}

X-Morph separates morphology information from task-specific correspondence information. The morphology specification describes the robot itself --- its kinematic tree, actuated joint order, joint limits, and nominal base height --- and is loaded directly from the robot description. The retargeting specification $\mathcal{C}^{h \rightarrow r}$ defines only the semantic body-part correspondences between source and target morphologies for a given task family. Because the morphology description is fixed, the same robot can be used with different task families simply by changing the retargeting specification.

\begin{table}[h!]
\centering
\small
\setlength{\tabcolsep}{6pt}
\caption{Retargeting specifications per robot and task family.
Source bodies refer to the G1 humanoid representation.}
\label{tab:retargeting_specs}
\begin{tabular}{llp{8.0cm}}
\toprule
Robot & Task & Body-part correspondences \\
\midrule
\multirow{2}{*}{Go2}
  & Locomotion
  & Human legs $\rightarrow$ 4 quadruped legs \\[2pt]
  & Loco+manip
  & Human legs $\rightarrow$ hind legs;\;
    human arms $\rightarrow$ front legs \\
\addlinespace
\multirow{2}{*}{Yuna}
  & Locomotion
  & Human legs $\rightarrow$ 6 hexapod legs \\[2pt]
  & Loco+manip
  & Human legs $\rightarrow$ 4 support legs;\;
    human arms $\rightarrow$ front 2 legs \\
\addlinespace
\multirow{2}{*}{B2-Z1}
  & Locomotion
  & Human legs $\rightarrow$ 4 quadruped legs \\[2pt]
  & Loco+manip
  & Human legs $\rightarrow$ 4 quadruped legs;\;
    human right arm $\rightarrow$ Z1 arm \\
\bottomrule
\end{tabular}
\end{table}

\begin{figure}[h!]
    \centering
    \includegraphics[width=0.95\linewidth]{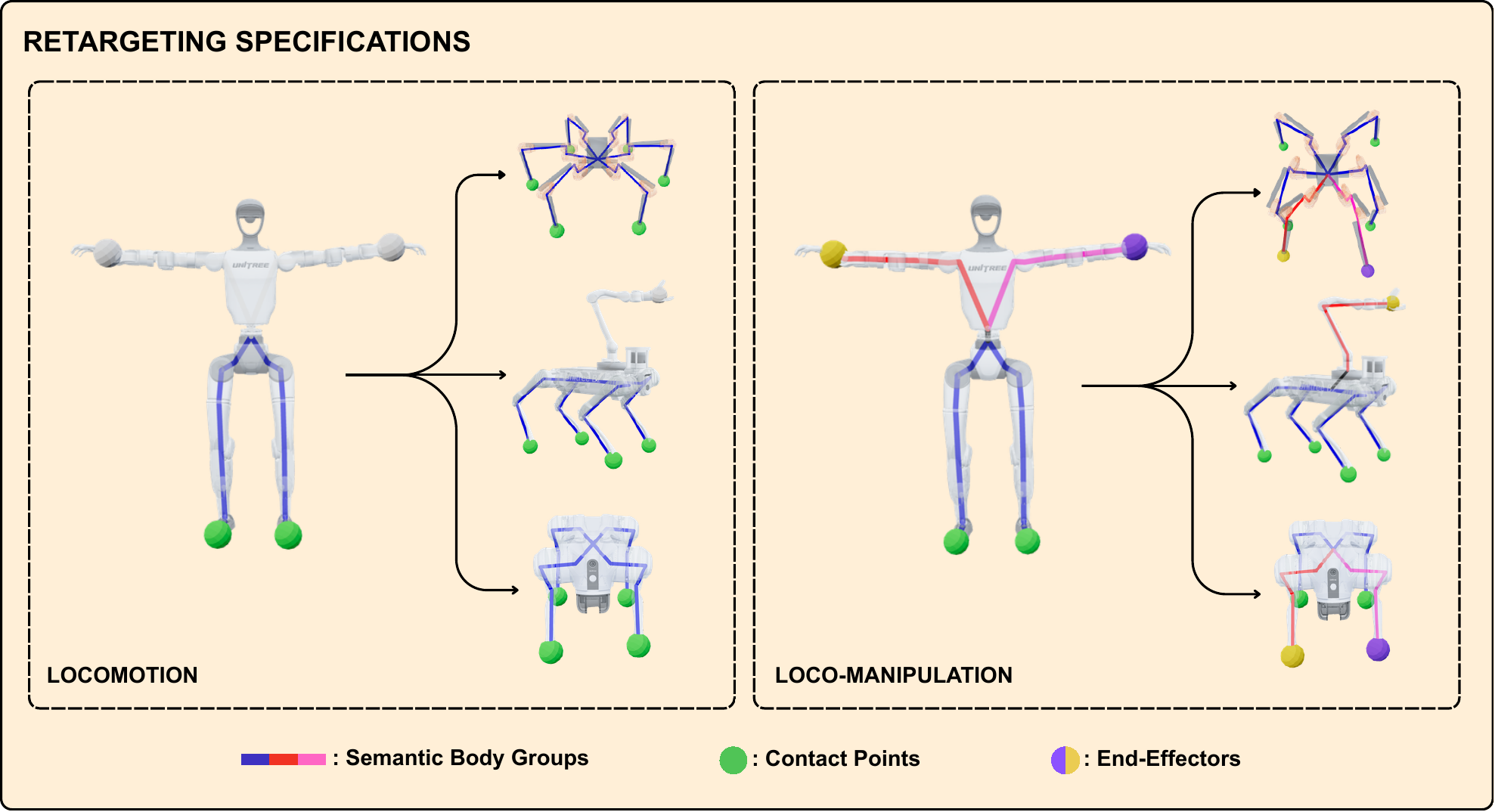}
    \caption{Retargeting specifications for locomotion and 
    loco-manipulation. Colored skeleton segments denote semantic 
    body groups used for cross-morphology correspondence, green 
    spheres denote contact points, and yellow/purple spheres denote 
    task end-effectors.}
    \label{fig:spec_graphic}
\end{figure}

In all experiments, source motions are first represented on the Unitree G1 humanoid. Table~\ref{tab:retargeting_specs} lists the body-part correspondences defined for each robot and task family. For manipulation tasks, end-effector correspondences additionally align task-relevant source and target end-effectors. Figure~\ref{fig:spec_graphic} illustrates representative mappings
across morphologies.

\section{Offline Retargeting Network}

The offline retargeting network follows the encode-transfer-decode structure of
PAN~\citep{Hu_2024}, but replaces dataset-specific skeleton assumptions with
robot morphology specifications and the local-frame motion representation
defined in the main paper. For each morphology $r$, the model contains a motion
encoder $E^r$, motion decoder $D^r$, skeleton encoder $S^r$, and discriminator
$\mathrm{Disc}^r$. The skeleton encoder maps robot morphology metadata, such as
the kinematic tree and joint offsets, into a morphology embedding $e^r$ used by
the motion decoder.

\begin{figure}[t]
    \centering
    \includegraphics[width=0.98\linewidth]{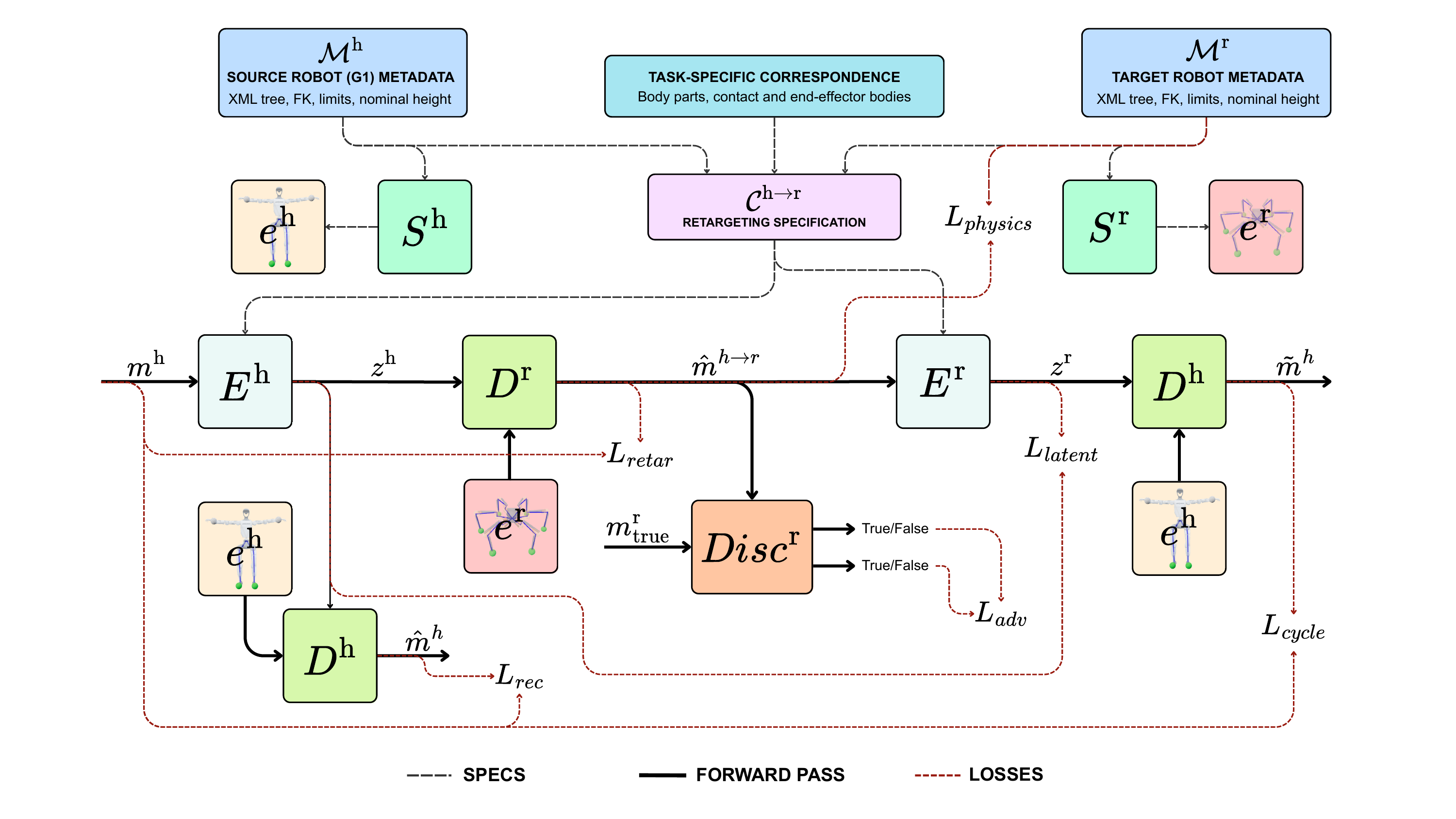}
    \caption{
    Offline retargeting teacher architecture. Solid arrows denote motion and
    latent flow, gray dashed arrows denote metadata or retargeting-specification
    conditioning, and red dashed arrows denote loss terms.
    }
    \label{fig:offline_retargeting_arch}
\end{figure}

Given a source G1 motion $m^h$, the source encoder produces a structured latent
sequence
\begin{equation}
    z^h = E^h(m^h), \qquad e^h = S^h(\mathcal{M}^h).
\end{equation}
The source decoder reconstructs the input motion using the source morphology
embedding,
\begin{equation}
    \hat{m}^h = D^h(z^h, e^h).
\end{equation}
To retarget the motion, the same latent sequence is decoded using the target
morphology embedding,
\begin{equation}
    \hat{m}^{h \rightarrow r} = D^r(z^h, e^r), \qquad e^r = S^r(\mathcal{M}^r).
\end{equation}
The retargeted motion is then encoded by the target encoder and decoded back
with the source decoder to impose cycle consistency,
\begin{equation}
    \tilde{m}^h =
    D^h\left(E^r(\hat{m}^{h \rightarrow r}), e^h\right).
\end{equation}
For clarity, Figure~\ref{fig:offline_retargeting_arch} shows only the
G1-to-target direction. During training, the same architecture is applied
symmetrically in the reverse direction, so both morphology-specific
autoencoders, discriminators, retargeting paths, and cycle paths are optimized
jointly.

The encoder uses learnable semantic body-part tokens, per-joint tokens, and
root-motion tokens. Joint positions are converted into per-joint tokens using
the scalar positional embedding used in PAN, which encodes both joint identity
within the body part and joint angle value. The root-motion features, namely
local linear velocity and yaw rate, are embedded separately with small MLPs and
appended as additional root-motion tokens. These tokens are processed by a
masked transformer encoder whose mask is derived from the retargeting
specification $\mathcal{C}^{h \rightarrow r}$, following PAN and additionally
allowing the extra root-motion tokens to attend to all other tokens. The
body-part token outputs are then passed through temporal body-part convolutions
to produce the latent sequence. The decoder mirrors this structure with temporal
deconvolution layers and separate heads for target joint positions and
root-motion features.

Training follows PAN's reconstruction, adversarial, retargeting, and
cycle-consistency objectives, with additional robot-specific losses computed on
the retargeted motion $\hat{m}^{h \rightarrow r}$. In particular, X-Morph adds
target robot FK, joint-limit, contact, grounding, and task end-effector losses
defined from the robot metadata and task-specific retargeting specification.
Relative to PAN, the key changes are: (i) morphology metadata, FK models, joint
limits, contact bodies, and end-effector bodies are loaded from robot
specifications; (ii) retargeting correspondences are task-specific and support
locomotion and loco-manipulation for the same morphology; (iii) root-motion
supervision uses separate local velocity and yaw-rate tokens; and (iv) the
learned retargeted motions are optimized for downstream trackability rather than
visual plausibility alone.

\section{Retargeting and Corrector Hyperparameters}

Tables~\ref{tab:teacher_loss_weights} and~\ref{tab:offline_arch} report the
offline retargeting architecture, optimization settings, and task-specific loss
weights. Tables~\ref{tab:causal_arch} and~\ref{tab:causal_losses} provide the causal retargeting student settings. 

The corrector model is a temporal residual network with 6 convolutional blocks, hidden dimension 192, kernel size 5, and dropout 0.1. Since this stage is offline, the network is non-causal and can use temporal context from the full motion. Losses described in Table~\ref{tab:corrector_losses} are computed from the corrected trajectory using robot forward kinematics, contact bodies, and joint limits. The resulting references are used for tracker training, student distillation, causal retargeting, and downstream deployment.

Finally, Table~\ref{tab:training_infra} reports representative training times for the major stages of the pipeline.

\begin{table}[h]
\centering
\small
\setlength{\tabcolsep}{6pt}
\renewcommand{\arraystretch}{1.05}
\caption{Offline retargeting network loss weights and key matching settings for
the main experiments. L+M denotes loco-manipulation.}
\label{tab:teacher_loss_weights}
\begin{tabular}{lcccc}
\toprule
Loss / setting & Go2 loco & Yuna loco & Yuna L+M & B2-Z1 L+M \\
\midrule
Reconstruction, $\lambda_{\mathrm{rec}}$
& 2.5 & 2.5 & 2.5 & 2.5 \\
Adversarial, $\lambda_{\mathrm{adv}}$
& 1.0 & 1.0 & 1.0 & 1.0 \\
Cycle latent, $\lambda_{\mathrm{cycle\_latent}}$
& 0.1 & 0.5 & 0.5 & 0.5 \\
Cycle FK, $\lambda_{\mathrm{cycle\_fk}}$
& 0.1 & 0.1 & 0.25 & 0.25 \\
Cycle motion, $\lambda_{\mathrm{cycle\_motion}}$
& 50.0 & 10.0 & 10.0 & 10.0 \\
Retarget velocity, $\lambda_{\mathrm{retar\_vel}}$
& 250.0 & 250.0 & 250.0 & 250.0 \\
Retarget yaw rate, $\lambda_{\mathrm{retar\_yaw}}$
& 25.0 & 25.0 & 25.0 & 20.0 \\
Yaw-rate scale
& 1.0 & 0.5 & 1.0 & 0.5 \\
Joint-limit loss
& 0.1 & 0.1 & 0.1 & 0.01 \\
Foot skating
& 1.0 & 1.0 & 5.0 & 0.1 \\
Grounding
& 10.0 & 5.0 & 5.0 & 5.0 \\
End-effector matching
& 0.0 & 0.0 & 0.5 & 0.5 \\
\bottomrule
\end{tabular}
\end{table}

\begin{table}[h]
\centering
\small

\begin{minipage}[t]{0.47\linewidth}
\centering
\caption{Causal retargeting network architecture and training settings.}
\label{tab:causal_arch}
\begin{tabular}{lc}
\toprule
Setting & Value \\
\midrule
Source history & 24 frames \\
Previous target context & 2 frames \\
Conv. channels & 128 \\
GRU hidden dim. & 256 \\
Conv. kernel & 3 \\
Batch size & 256 \\
Learning rate & $5 \times 10^{-4}$ \\
Weight decay & $10^{-4}$ \\
\bottomrule
\end{tabular}
\end{minipage}
\hfill
\begin{minipage}[t]{0.47\linewidth}
\centering
\caption{Causal retargeting network loss weights.}
\label{tab:causal_losses}
\begin{tabular}{lc}
\toprule
Loss term & Weight \\
\midrule
Joint imitation & 5.0 \\
Root-motion consistency & 1.0 \\
Temporal smoothness & 0.5 \\
Joint limits & 0.01 \\
\bottomrule
\end{tabular}
\end{minipage}

\end{table}

\begin{table}[h]
\centering
\small

\begin{minipage}[t]{0.49\linewidth}
\centering
\caption{Offline retargeting Network architecture and optimization settings.}
\label{tab:offline_arch}
\begin{tabular}{ll}
\toprule
Component & Setting \\
\midrule
Transformer layers & 1 \\
Token latent dim. & 32 \\
Temporal conv/deconv & 2 / 2 \\
Temporal kernel & 15 \\
Batch size & 128 \\
Training epochs & 3001 \\
Generator LR & $10^{-4}$ \\
Discriminator LR & $10^{-4}$ \\
\bottomrule
\end{tabular}
\end{minipage}
\hfill
\begin{minipage}[t]{0.46\linewidth}
\centering
\caption{Default physics corrector loss weights.}
\label{tab:corrector_losses}
\begin{tabular}{lc}
\toprule
Loss term & Weight \\
\midrule
Joint preservation & 10.0 \\
Root vel. preservation, $xy$ & 0.5 \\
Root vel. preservation, $z$ & 1.0 \\
Yaw-rate preservation & 0.5 \\
Residual smoothness & 0.05 \\
Joint velocity smoothness & 0.05 \\
Joint acceleration smoothness & 0.05 \\
Foot skating & 1.0 \\
Grounding & 25.0 \\
Joint limits & 0.01 \\
\bottomrule
\end{tabular}
\end{minipage}

\end{table}

\begin{table}[h]
\centering
\small
\caption{Training hardware and approximate wall-clock time for the main
X-Morph components.}
\label{tab:training_infra}
\begin{tabular}{lccc}
\toprule
Component & GPU & Time & Output \\
\midrule
Offline retargeting Network  & RTX 3090 & 2 h & Raw robot references \\
Physics corrector & RTX 3090 & 5 min & Cleaned robot references \\
Causal retargeting Network & RTX 3090 & 15 min & Online retargeting model \\
Privileged tracker teacher & RTX 4090 & 2.5 h & Privileged tracking policy \\
Tracker student & RTX 4090 & 2 h & Deployable tracking policy \\
\bottomrule
\end{tabular}
\end{table}

\section{Tracker Rewards and Domain Randomization}

The tracker converts cleaned robot references into closed-loop control policies.
For each target morphology, we train a reference-conditioned policy that observes
the robot proprioceptive state together with a short reference-motion preview.
At control step $t$, the deployable policy receives the current robot state
$s_t$ and reference features from the target trajectory. For a reference-frame
offset set $\mathcal{H}$, the reference observation is
\[
    r_t =
    \left\{
    \left[
    q^r_{t+k},\;
    v^r_{b,t+k},\;
    \omega^r_{b,t+k},\;
    R^r_{b,t+k}
    \right]
    \right\}_{k \in \mathcal{H}},
\]
where $q^r$ denotes reference joint positions, $v^r_b$ and $\omega^r_b$
denote reference base linear and angular velocity, and $R^r_b$ denotes the
reference base orientation. The policy outputs target joint positions for the
low-level PD controller.

We use an APEX-style \citep{sood2025apexactionpriorsenable, 10802000}  decaying action prior during tracker training. For a current robot joint position $q_t$, a reference joint position at the current motion frame $\hat q_t$, and the joint-position action predicted by the policy $a^\pi_t$ , the executed joint-position command is 
\[
    a_t = a^\pi_t + \lambda_t(\hat q_t - q_t),
\]
where $\lambda_t$ decays from 1 to 0 during training. Early in training, this
prior biases exploration toward reference-like states and makes imitation
rewards easier to discover. As $\lambda_t$ decays, the learned policy must
produce the tracking behavior on its own. At convergence and deployment, the
policy no longer relies on this action prior.

Tracker training follows a privileged-teacher and deployable-student structure.
The privileged teacher receives clean full-state observations and a larger
temporal reference window, including both past and future reference frames. The deployable student receives only deployment-available proprioception and a short reference preview. For Yuna, the policy is conditioned on a one-step reference preview, i.e. the current and next reference frames. For B2-Z1, we use the same one-step reference preview, but provide the policy with a flattened history of five recent observations. Thus, the B2-Z1 policy does not access a longer future horizon; instead, it has an additional causal context over recent proprioceptive states, actions, and reference commands. This is useful for loco-manipulation, where end-effector tracking, base stabilization, and contact timing are temporally coupled.

\begin{table}[h]
\centering
\small

\begin{minipage}[t]{0.49\linewidth}
\centering
\caption{Proprioceptive observation noise randomization for deployable tracker training.}
\label{tab:tracker_proprio_noise}
\begin{tabular}{lc}
\toprule
Observation term & Range \\
\midrule
Base angular velocity & $[-0.8,0.8]$ \\
Projected gravity & $[-0.12,0.12]$ \\
Joint position & $[-0.03,0.03]$ \\
Joint velocity & $[-3.0,3.0]$ \\
\bottomrule
\end{tabular}

\vspace{2em}

\caption{Yuna tracker domain randomization.}
\label{tab:yuna_tracker_dr}
\begin{tabular}{ll}
\toprule
Quantity & Range \\
\midrule
Static friction & $[0.3,1.0]$ \\
Dynamic friction & $[0.3,0.8]$ \\
Base mass offset & $[-1.0,2.0]$ kg \\
Non-base link mass scale & $[0.8,1.2]$ \\
COM offset $x$ & $[-0.05,0.05]$ m \\
COM offset $y$ & $[-0.05,0.05]$ m \\
COM offset $z$ & $[-0.05,0.05]$ m \\
Stiffness scale & $[0.8,1.2]$ \\
Damping scale & $[0.8,1.2]$ \\
Push interval & $[8,12]$ s \\
Push vel. $x$ & $[-0.5,0.5]$ m/s \\
Push vel. $y$ & $[-0.5,0.5]$ m/s \\
Push roll vel. & $[-0.4,0.4]$ rad/s \\
Push pitch vel. & $[-0.4,0.4]$ rad/s \\
Push yaw vel. & $[-0.4,0.4]$ rad/s \\
\bottomrule
\end{tabular}
\end{minipage}
\hfill
\begin{minipage}[t]{0.49\linewidth}
\centering
\caption{Go2 tracker domain randomization.}
\label{tab:go2_tracker_dr}
\begin{tabular}{ll}
\toprule
Quantity & Range \\
\midrule
Stiffness scale & $[0.9,1.1]$ \\
Damping scale & $[0.9,1.1]$ \\
Push interval & $[10,15]$ s \\
Push vel. $x$ & $[-0.2,0.2]$ m/s \\
Push vel. $y$ & $[-0.2,0.2]$ m/s \\
Push roll vel. & $[-0.15,0.15]$ rad/s \\
Push pitch vel. & $[-0.15,0.15]$ rad/s \\
Push yaw vel. & $[-0.2,0.2]$ rad/s \\
\bottomrule
\end{tabular}

\vspace{2em}

\caption{B2-Z1 tracker domain randomization.}
\label{tab:b2_z1_tracker_dr}
\begin{tabular}{ll}
\toprule
Quantity & Range \\
\midrule
Base mass scale & $[0.9,1.15]$ \\
Non-base link mass scale & $[0.85,1.15]$ \\
COM offset $x$ & $[-0.04,0.04]$ m \\
COM offset $y$ & $[-0.03,0.03]$ m \\
COM offset $z$ & $[-0.03,0.03]$ m \\
External force & $[-40,40]$ N \\
External torque & $[-20,20]$ Nm \\
Z1 mount $x$ & $[-0.003,0.003]$ m \\
Z1 mount $z$ & $[-0.002,0.002]$ m \\
Stiffness scale & $[0.85,1.15]$ \\
Damping scale & $[0.85,1.15]$ \\
\bottomrule
\end{tabular}
\end{minipage}

\end{table}

\begin{table}[h]
\centering
\scriptsize
\setlength{\tabcolsep}{4pt}
\caption{Tracker reward weights across robot platforms.}
\label{tab:tracker_reward_weights}
\begin{tabular}{lccc}
\toprule
Reward term & Go2 & Yuna & B2-Z1 \\
\midrule
Joint pos. imitation & 1.0 & 1.0 & -- \\
Leg joint pos. imitation & -- & -- & 0.5 \\
Arm joint pos. imitation & -- & -- & 2.0 \\
Projected gravity imitation & 0.5 & 0.5 & 0.5 \\
Foot pos. imitation & 1.0 & -- & 1.0 \\
End-effector pos. imitation & -- & 2.0 & 2.0 \\
World foot pos. imitation & 0.75 & -- & -- \\
World base pos. imitation & 0.5 & -- & -- \\
Cmd. lin. vel., $xy$ & 1.5 & 1.5 & 2.0 \\
Cmd. lin. vel., $z$ & 0.5 & -- & 0.75 \\
Cmd. yaw rate & 1.0 & 1.0 & 0.5 \\
Base height error & $-10.0$ & $-10.0$ & $-10.0$ \\
Foot slip & $-0.08$ & $-0.08$ & $-0.08$ \\
Impact penalty & $-5{\times}10^{-3}$ & $-5{\times}10^{-3}$ & $-5{\times}10^{-3}$ \\
Airborne contact & $-0.75$ & -- & -- \\
Undesired contacts & $-1.0$ & $-1.0$ & $-3.0$ \\
Base ang. vel., $xy$ & $-0.05$ & $-0.05$ & $-0.05$ \\
Joint acceleration & $-2.5{\times}10^{-7}$ & $-2.5{\times}10^{-7}$ & -- \\
Leg joint acceleration & -- & -- & $-2.5{\times}10^{-7}$ \\
Arm joint acceleration & -- & -- & $-5.0{\times}10^{-7}$ \\
Joint torque & $-10^{-5}$ & $-10^{-5}$ & -- \\
Leg joint torque & -- & -- & $-10^{-5}$ \\
Arm joint torque & -- & -- & $-2.0{\times}10^{-5}$ \\
Action rate & $-10^{-2}$ & $-2.0{\times}10^{-2}$ & -- \\
Leg action rate & -- & -- & $-10^{-2}$ \\
Arm action rate & -- & -- & $-2.5{\times}10^{-2}$ \\
Action smoothness & $-10^{-2}$ & $-5.0{\times}10^{-2}$ & -- \\
Leg action smoothness & -- & -- & $-10^{-2}$ \\
Arm action smoothness & -- & -- & $-2.5{\times}10^{-2}$ \\
Joint limits, legs & -- & -- & $-2.0$ \\
Default leg pose & -- & -- & $-0.25{\times}10^{-3}$ \\
Leg distance & -- & -- & $-1.5$ \\
Foot contact force & -- & -- & $-10^{-4}$ \\
\bottomrule
\end{tabular}
\end{table}

\clearpage
\section{Baseline and Related Work Discussion}
Conducting a qualitative comparison with other cross-morphology retargeting methods
is difficult because their assumptions, settings, and scope vary substantially. Further, few of these works have concrete baselines and openly released code. While the majority of these works focus on offline reference generation or deployment settings, X-Morph additionally trains deployable
tracking policies and supports real-time reference generation from video or
text-conditioned motion sources. We therefore attempt to broadly summarize the main methodological differences between prior work and our approach in Table~\ref{tab:baseline_comparison}.

\begin{table}[h!]
\centering
\small
\setlength{\tabcolsep}{5pt}
\renewcommand{\arraystretch}{1.05}
\caption{Comparison with cross-morphology motion transfer methods.}
\label{tab:baseline_comparison}
\begin{tabular}{lp{7.0cm}ccc}
\toprule
Method & Scope & Hardware & Interactive \\
\midrule
PAN~\citep{Hu_2024}
& broad offline character retargeting
& -- & --  \\
ACE~\citep{li2023ace}
& designed embodiment/task mappings
& \checkmark & -- \\
HumanConQuad~\citep{kim2022humanconquad}
& quadruped teleop with predefined robot motion experts
& \checkmark & \checkmark \\
MoReFlow~\citep{kim2025moreflowmotionretargetinglearning}
& offline transfer for a specified source-target pair
& -- & -- \\
CrossLoco~\citep{li2024crossloco}
& quadruped locomotion transfer and tracking
& -- & -- \\
\midrule
X-Morph
& reusable task-family models for multiple legged robots
& \checkmark & \checkmark \\
\bottomrule
\end{tabular}
\end{table}

Other cross-embodiment methods often rely on more rigid correspondence structures or narrower task settings. ACE, for example, uses structured correspondences between source and target embodiments, but is primarily designed around specified embodiment mappings and does not demonstrate the same combination of unseen human motion generalization, multi-robot deployment, and interactive real-time control. Quadruped-focused human retargeting methods similarly tend to target a single robot morphology or a restricted motion class. In contrast, X-Morph separates morphology specifications from task-specific retargeting specifications, allowing the same pipeline to support quadrupeds, hexapods, and quadruped-manipulators. While X-Morph also trains separate models for each source-target robot pair, the models are trained on broad motion collections and reused across motions within a task family.

Locomotion tracking methods such as CrossLoco produce deployable policies, but they are designed primarily for locomotion and do not solve the full problem of converting diverse human motion sources into reusable references for multiple non-humanoid morphologies. X-Morph is complementary to these works: it uses a learned tracker for execution, but focuses on producing the cross-morphology references and causal retargeting models needed to turn human motion priors into robot behavior across platforms.

\section{Generalization to Unseen Motions}
\begin{figure}[h]
    \centering
    \begin{subfigure}[t]{0.48\linewidth}
        \centering
        \includegraphics[width=\linewidth]{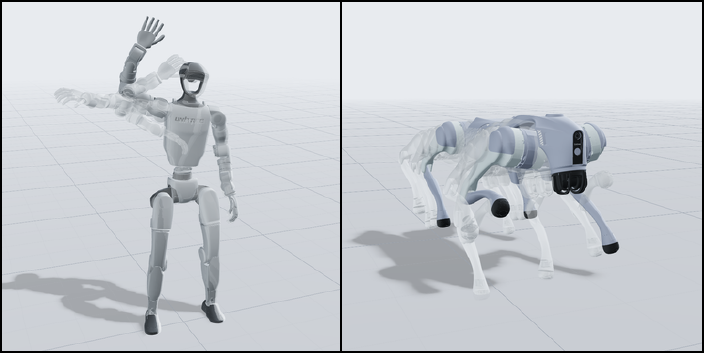}\
        \caption{Go2 right-paw manipulation from a right-arm source motion,
        despite training only on left-paw manipulation data.}
    \end{subfigure}
    \hfill
    \begin{subfigure}[t]{0.48\linewidth}
        \centering
        \includegraphics[width=\linewidth]{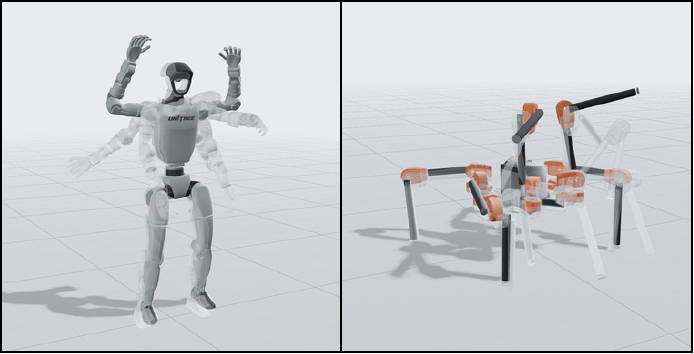}
        \caption{Yuna front-leg manipulation with larger range and diversity
        than the limited target manipulation demonstrations.}
    \end{subfigure}
    \caption{
    Qualitative generalization to unseen manipulation motions. X-Morph transfers
    source motion intent beyond the exact target demonstrations used during
    retargeting training.
    }
    \label{fig:unseen_generalization}
\end{figure}

We qualitatively evaluate whether X-Morph can retarget motions outside the
narrow demonstrations available for a target morphology. Figure~\ref{fig:unseen_generalization}
shows two representative cases. First, the Go2 manipulation model is trained
only with left-paw manipulation references, yet it can retarget a standing
human right-arm motion into a plausible right-paw manipulation behavior. This
suggests that the learned correspondence structure is not simply memorizing a
single limb trajectory, but can reuse the source motion representation across
symmetric target limbs.

Second, the Yuna loco-manipulation model is trained with limited examples of
front-leg manipulator motion. Despite this, it generalizes to a wider range of
human upper-body manipulation motions, producing front-leg motions with larger
range and more diverse spatial structure than those present in the target
training set.

\section{Video Teleoperation}

In the full local evaluation setup, the video pipeline produced references at
24.3 Hz without visualization and 20.0 Hz with visualization. When deployed on
hardware without the concurrent sim2sim evaluation overhead, the same pipeline
reached a peak reference publishing rate of 28.9 Hz with a 30 FPS camera stream.

\begin{table}[h]
\centering
\small
\caption{Per-stage latency for the live video-to-reference pipeline evaluated with sim2sim. Values are mean latencies after removing warmup and shutdown rows.}
\label{tab:video_latency}
\begin{tabular}{lcc}
\toprule
Stage & No visualization & With visualization \\
& (ms) & (ms) \\
\midrule
Camera/read & 0.2 & 0.1 \\
FastSAM3D Body & 29.2 & 30.3 \\
MHR to SMPL & 3.0 & 2.9 \\
SMPL FK & 4.2 & 4.4 \\
GMR retargeting & 2.1 & 2.2 \\
Causal retargeting student & 1.3 & 1.3 \\
ZMQ publish & 0.1 & 0.2 \\
Viewer & 0.0 & 7.5 \\
\midrule
Total & 41.1 & 49.9 \\
Effective reference rate & 24.3 Hz & 20.0 Hz \\
\bottomrule
\end{tabular}
\end{table}

\begin{table}[h]
\centering
\small
\caption{Performance analysis of the live video-to-reference pipeline.}
\label{tab:video_runtime}
\begin{tabular}{lccc}
\toprule
Setting & Visualization & Mean ref. rate & Peak ref. rate \\
\midrule
Local eval + sim2sim logging & Off & 24.3 Hz & 26.8 Hz \\
Local eval + sim2sim logging & On & 20.0 Hz & 21.2 Hz \\
Hardware deployment & Off & 28.1 Hz & 28.9 Hz \\
\bottomrule
\end{tabular}
\end{table}

\end{document}